\documentclass[10pt,twocolumn,letterpaper]{article}
\usepackage[accsupp]{axessibility}
\usepackage{wacv}              

\usepackage{graphicx}
\usepackage{amsmath}
\usepackage{amssymb}
\usepackage{booktabs}
\usepackage{mathtools}
\usepackage{amsthm}
\usepackage[noend]{algpseudocode}
\usepackage{algorithm}
\usepackage{bm}
\usepackage[symbol]{footmisc}

\usepackage{tabularx}
\usepackage{calc}
\usepackage{amstext}
\usepackage{pifont}
\usepackage{siunitx}
\usepackage[table,xcdraw]{xcolor}
\usepackage[export]{adjustbox}
\usepackage{multirow}
\usepackage{lineno}
\newcommand{\cmark}{\ding{51}}%
\newcommand{\xmark}{\ding{55}}%

\usepackage{wrapfig}

\usepackage[pagebackref,breaklinks,colorlinks]{hyperref}

\usepackage[capitalize]{cleveref}
\crefname{section}{Sec.}{Secs.}
\Crefname{section}{Section}{Sections}
\Crefname{table}{Table}{Tables}
\crefname{table}{Tab.}{Tabs.}

\def\wacvPaperID{1404} 
\def\confName{WACV}
\def\confYear{2024}

\makeatletter
\newcommand{\printfnsymbol}[1]{%
  \textsuperscript{\@fnsymbol{#1}}%
}

\begin{document}

\title{Continual Learning of Unsupervised Monocular Depth from Videos}

\author{Hemang Chawla\textsuperscript{1,}\thanks{Equal contribution. ~~~\textsuperscript{$\dagger$}Equal advisory role.},
Arnav Varma\textsuperscript{2,}\printfnsymbol{1},
Elahe Arani\textsuperscript{1,3,$\dagger$}, and
Bahram Zonooz\textsuperscript{1,2,$\dagger$} \\
  \textsuperscript{1}Eindhoven University of Technology (TU/e)~~~
  \textsuperscript{2}TomTom~~~
  \textsuperscript{3}Wayve~~~\\
{\tt\small \{h.chawla, e.arani, b.zonooz\}@tue.nl, arnav.varma@tomtom.com}}
\maketitle

\begin{abstract}
Spatial scene understanding, including monocular depth estimation, is an important problem in various applications such as robotics and autonomous driving. While improvements in unsupervised monocular depth estimation have potentially allowed models to be trained on diverse crowdsourced videos, this remains underexplored as most methods utilize the standard training protocol wherein the models are trained from scratch on all data after new data is collected. Instead, continual training of models on sequentially collected data would significantly reduce computational and memory costs. Nevertheless, naive continual training leads to catastrophic forgetting, where the model performance deteriorates on older domains as it learns on newer domains, highlighting the trade-off between model stability and plasticity. 
While several techniques have been proposed to address this issue in image classification, the high-dimensional and spatiotemporally correlated outputs of depth estimation make it a distinct challenge. 
To the best of our knowledge, no framework or method currently exists focusing on the problem of continual learning in depth estimation. 
Thus, we introduce a framework that captures the challenges of continual unsupervised depth estimation (\textit{CUDE}), and define the necessary metrics to evaluate model performance. 
We propose a rehearsal-based dual-memory method \textit{MonoDepthCL}, which utilizes spatiotemporal consistency for continual learning in depth estimation, even when the camera intrinsics are unknown.\footnote[4]{\href{https://github.com/NeurAI-Lab/CUDE-MonoDepthCL.git}{https://github.com/NeurAI-Lab/CUDE-MonoDepthCL.git}}
\end{abstract}

\section{Introduction}
\label{sec:introduction}

\begin{figure}[t]
\centering
\includegraphics[width=0.49\textwidth]{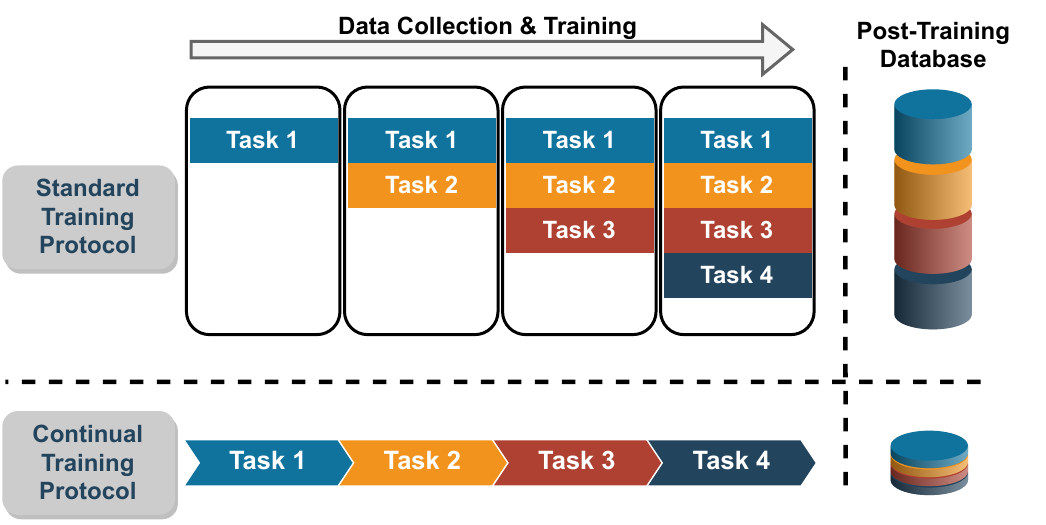}
\includegraphics[width=0.48\textwidth]{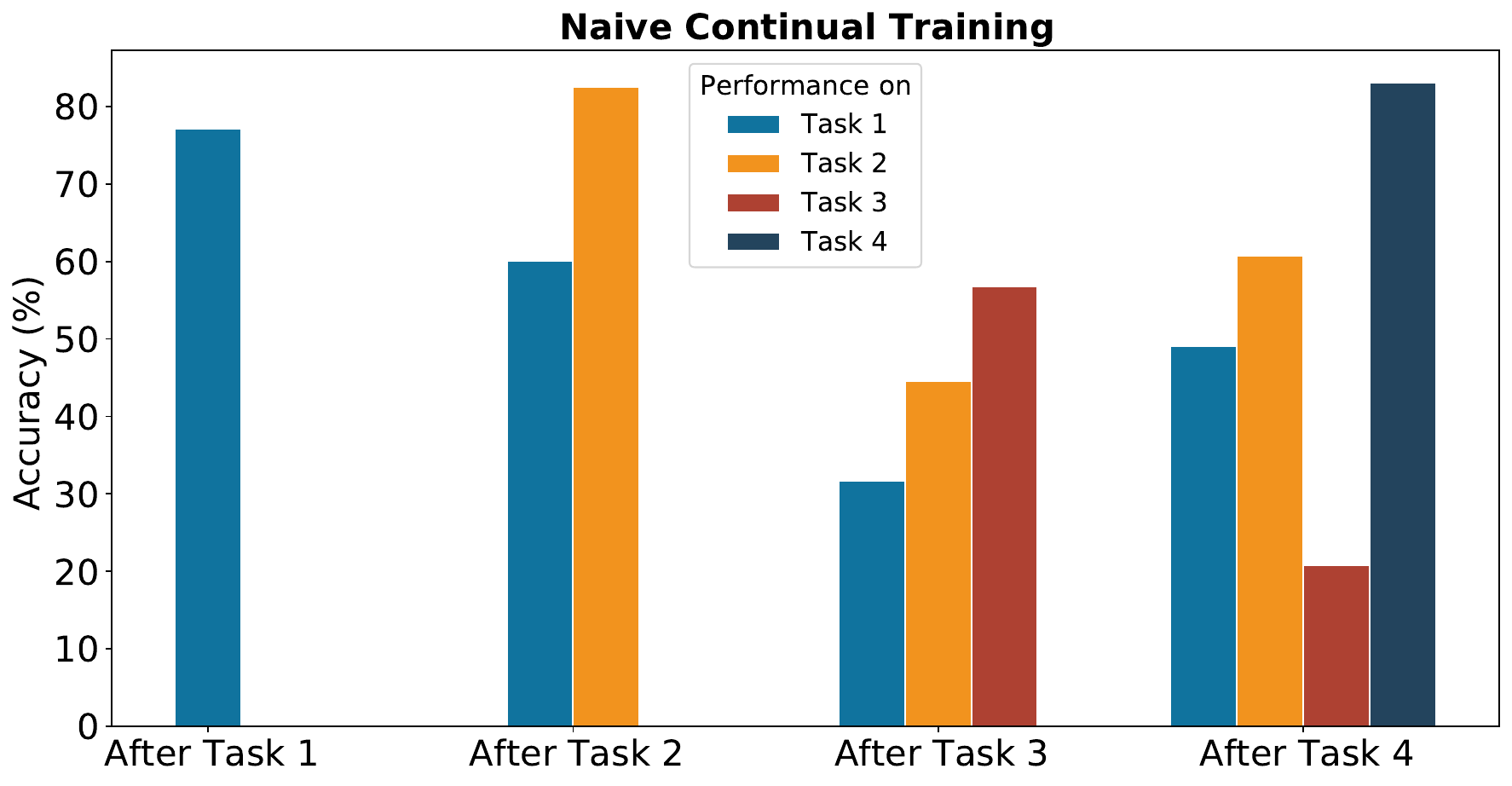}
\caption{(Top) Standard Training Protocol involves repeatedly training the model from scratch as new data is collected. All collected data is required for future training. The Continual Training Protocol sequentially trains the model as new data is collected. Only a limited amount of data is generally stored when training continually. (Bottom) Naive continual training on the introduced CUDE framework leads to catastrophic forgetting where the depth estimation accuracy~\cite{ladicky2014pulling} on a task is highest after training on that task but is significantly reduced after training on future tasks.}
\label{fig:problem}
\end{figure}

Vision-based systems have improved tremendously over the years with better semantic and spatial scene understanding capabilities~\cite{hu2020probabilistic}.
Particularly, the capability to estimate scene depth has found applications in augmented reality, autonomous navigation, object-grasping, robot-assisted surgery, and more~\cite{yang2020d3vo, kalia2019real, guizilini2021sparse, ming2021deep}. With the increasing demand for cost-effective, lightweight, and flexible depth estimation solutions, monocular camera-based methods have emerged as a promising alternative to the use of LIDARs or time-of-flight sensors. Moreover, unlike traditional approaches that rely on hand-crafted features from multiple views~\cite{schonberger2016structure}, deep learning has enabled depth estimation from single images~\cite{lee2021patch,patil2022p3depth, guizilini20203d}. Among these, unsupervised methods that do not require ground-truth labels are often preferred over supervised methods given the associated data-collection costs. These unsupervised methods potentially allow the training of depth estimation models on crowdsourced videos with unknown camera intrinsics~\cite{gordon2019depth}. Nevertheless, it remains an underexplored problem due to the challenges associated with the standard training protocol as shown in Figure~\ref{fig:problem}. Within the standard protocol, the models are retrained from scratch on all the available data after new data is collected, thus incurring high computational time and energy costs. Furthermore, to ensure that the models can be updated in the future, it is necessary to continue storing all the collected data post-training.

Instead, the continual training protocol involves incremental training of the model in sequential or stream data from different domains~\cite{lesort2020continual}. 
Continual Learning (CL) is also necessitated for practical deployments of depth estimation networks, such as in the case of a self-driving car or robot navigating through various locations, or when adapting models pre-trained on simulated data to real environments~\cite{rusu2017sim}. 
Additionally, privacy concerns may be necessary in scenarios where the robot does not have permission to store data on a server to train and update its models~\cite{butler2015privacy}, and has limited on-board memory.
However, adapting the model for the new domain is not sufficient, as it must maintain its depth estimation capability on earlier domains as well. 
Therefore, there is a fundamental dilemma between maintaining the stability of the old information and the plasticity to adapt to new information for CL~\cite{abraham2005memory,lesort2020continual}. Naive CL without a strategy to handle this dilemma will result in catastrophic forgetting~\cite{mccloskey1989catastrophic}, as can be seen in Figure~\ref{fig:problem}.

 CL methods such as regularization~\cite{kirkpatrick2017overcoming,schwarz2018progress,zenke2017continual}, parameter isolation~\cite{yoon2017lifelong,adel2019continual,kang2022forget}, and rehearsal~\cite{DBLP:conf/iclr/RiemerCALRTT19,buzzega2020dark,boschini2022class} have been utilized to address this dilemma for image classification. 
 However, depth estimation is a distinct challenge because of its high-dimensional output space. Additionally, it is typically required to make highly correlated predictions through the use of local spatial relations within images and temporal relations within videos, where the depth at one pixel may depend on the depth at other pixels. 
 Thus, there is a need for a framework that includes a set of sequential tasks that represent these challenges, as well as suitable metrics for evaluating the performance of the models. Thereafter, methods can be designed to consider spatiotemporal relations while addressing the stability-plasticity dilemma specifically for CL in unsupervised monocular depth estimation.

Hence, we introduce a framework for continual unsupervised depth estimation (\textit{CUDE}) to overcome the aforementioned gap. It consists of a setup of four tasks, each on a different dataset with unique characteristics representative of the challenges of domain and depth range shifts across simulated or real and indoor or outdoor scenes. Additionally, we provide the appropriate error and accuracy metrics that can be used to evaluate the performance of depth estimation trained within the continual training protocol. Finally, we propose a dual-memory rehearsal-based method \textit{MonoDepthCL} with a spatiotemporal consistency loss for CL in depth estimation.
We showcase how sequential unsupervised learning of monocular depth across multiple tasks enables the development of spatial scene understanding, even when the camera intrinsics may be unknown. 
Our contributions are as follows:
\begin{itemize}
    \item We develop a framework for benchmarking of continual learning methods for unsupervised depth estimation under real-world ever-changing scenarios such as different cameras, diverse weather and lighting conditions, disparate depth ranges, and sim-to-real, indoor-to-outdoor, and outdoor-to-indoor domain shifts. 
    \item We define metrics to evaluate the continual learning methods in the framework such that they capture various aspects of continual learning performance such as final performance, performance across the learning trajectory, and stability-plasticity trade-off. 
    \item We propose a method - \textit{MonoDepthCL}, for continual learning of unsupervised monocular depth estimation using multiple models to explicitly capture stability and plasticity separately. Aided by a novel spatiotemporal consistency loss, \textit{MonoDepthCL} proves to be effective for continual learning and dealing with the stability-plasticity trade-off. \textit{MonoDepthCL} is also shown to be effective even when the camera intrinsics are unknown.
\end{itemize}

\section{Related Works}
\label{sec:related_works}

\begin{figure*}[t]
\centering
\includegraphics[width=\textwidth]{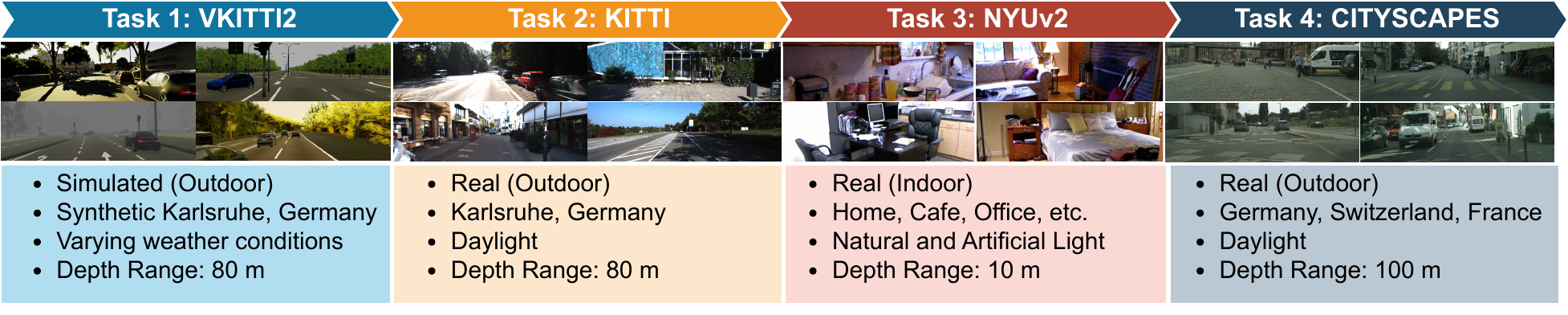}
\caption{Continual Unsupervised Depth Estimation (CUDE) framework with four sequential tasks operating in diverse environments, weather and lighting conditions, and depth ranges.}
\label{image:framework}
\end{figure*}

\subsection{Monocular Unsupervised Depth Estimation}
\label{subsec:rel_work_monodepth}

Monocular depth estimation is considered an important computer vision task for many applications.
With the advent of deep learning, several supervised~\cite{eigen2014depth,ranftl2021vision}, as well as unsupervised~\cite{godard2019digging,bian2021auto} approaches for depth estimation have been proposed. 
Further research has improved the depth estimation results in independent and identically distributed (i.i.d.) training using multiple
modalities~\cite{chawla2021multimodal,guizilini20203d}, 
newer architectures~\cite{DBLP:conf/visapp/VarmaCZA22,guizilini2022multi}, 
advances in feature extraction~\cite{lyu2021hr,shu2020feature}, and 3D geometry~\cite{bian2019unsupervised,Wang_2021_ICCV}. Nevertheless, estimating dense depths from monocular images for continually shifting distributions, where the previous data become unavailable as in the real world, remains an understudied problem.

\subsection{Continual Learning for Dense Prediction}
Most of the CL research has focused on classification tasks, with limited attention to dense prediction tasks. Early works on CL in image classification focused on regularizing important parameters for previous tasks \cite{kirkpatrick2017overcoming,schwarz2018progress,zenke2017continual}, or complete or partial isolation of parameters relevant to each task \cite{yoon2017lifelong,adel2019continual,kang2022forget}. However, these approaches generally undergo catastrophic forgetting without task identification at test time~\cite{farquhar_towards_2018}. Rehearsal-based methods, such as experience replay (ER)~\cite{DBLP:conf/iclr/RiemerCALRTT19} resolve this issue by retraining on old data stored in a small memory buffer updated with replacement~\cite{buzzega2020dark,boschini2022class,pham2021dualnet,arani2022learning}. 
Nevertheless, it is non-trivial to extend these methods to continual learning for dense prediction tasks. 
Although some methods have been developed that focus on continual semantic segmentation~\cite{douillard2021plop,maracani2021recall}, they continue to address the challenge posed by the addition of newer class labels. 
Recently, some works have focused on related issues of domain adaptation for the spatial task of depth estimation~\cite{kuznietsov2021comoda,zhang2020online}. These methods rarely focus on the issue of catastrophic forgetting or are limited to a one-step transfer of learned knowledge to a single new environment. 
Voedisch \etal \cite{voedisch2022continual} focuses on the related task of odometry.
However, it uses an infinite buffer, giving it access to all previously seen data at all times, additionally causing high memory expense and privacy concerns.  
To the best of our knowledge, no existing work deals with the challenges of CL in dense unsupervised depth estimation. Therefore, we introduce a framework for comprehensive evaluation of CL methods for unsupervised depth estimation, and develop a rehearsal-based method for the same.

\section{Framework}
\label{sec:Framework}

Since depth estimation is a regression method with no distinct classes, CL for depth estimation is akin to a domain-incremental learning scenario, in which the input distribution changes as training progresses. This is reflected in Figure~\ref{fig:problem} where the continual training protocol is different from standard training, where the model is trained on all available datasets simultaneously. We, thus, define the CUDE framework with four tasks, each corresponding to an individual dataset as shown in Figure~\ref{image:framework}. Each task corresponds to training the depth on a unique set of videos captured in varying environments from different cameras, under diverse weather and lighting conditions, and capturing disparate depth ranges to mimic real-world ever-changing scenarios. 
The task sequence is VKITTI2~\cite{cabon2020vkitti2} $\rightarrow$ KITTI~\cite{geiger2013vision} $\rightarrow$ NYUv2~\cite{silberman2012indoor} $\rightarrow$ Cityscapes~\cite{cordts2016cityscapes}. Here, VKITTI2 is a virtual photo-realistic dataset generated using the Unity game engine for the simulated urban setting of Karlsruhe, Germany in various imaging and weather conditions. KITTI is an outdoor scenario dataset captured in Karlsruhe, Germany, consisting of challenging scenes from both urban and highway scenarios with ground truth depth measured from a LiDAR. NYUv2 is an indoor dataset consisting of images and the corresponding ground truth depth captured from a Kinect RGBD camera. Cityscapes is another outdoor vision dataset captured in multiple locations in Germany, France, and Switzerland with ground truth depth measured using stereo vision.

This task order also spans complex domain shifts such as indoor-to-outdoor, outdoor-to-indoor, and sim-to-real. Though there could have been $24$ possible task orders, we eliminate those sequences where the simulated dataset would be in the middle or end of the sequence, as they are not applicable to real-world scenarios, which typically have deployment in the real-world as the target. Additionally, $4$ of the remaining $6$ permutations lack either the indoor-to-outdoor or outdoor-to-indoor domain shift. Finally, from the remaining $2$ sequences, the selected sequence allows us to demonstrate the impact of transitioning from sim-to-real for the same scene and camera setup, which is more realistic as robots are often trained first on the closest possible simulated version of the targeted environment before obtaining data for the real-world environment. 
Such a training sequence could be utilized to train perception systems for robots that operate both indoors and outdoors, such as security robots, hygiene robots, assistance robots, etc. By capturing domain shifts from simulated to real environments, outdoor to indoor environments, and vice versa, our setup aims to examine both the forgetfulness and adaptability of the CL models across different camera setups, scene distributions, and depth ranges.  

\begin{figure*}[t]
\centering
\includegraphics[width=0.975\textwidth]{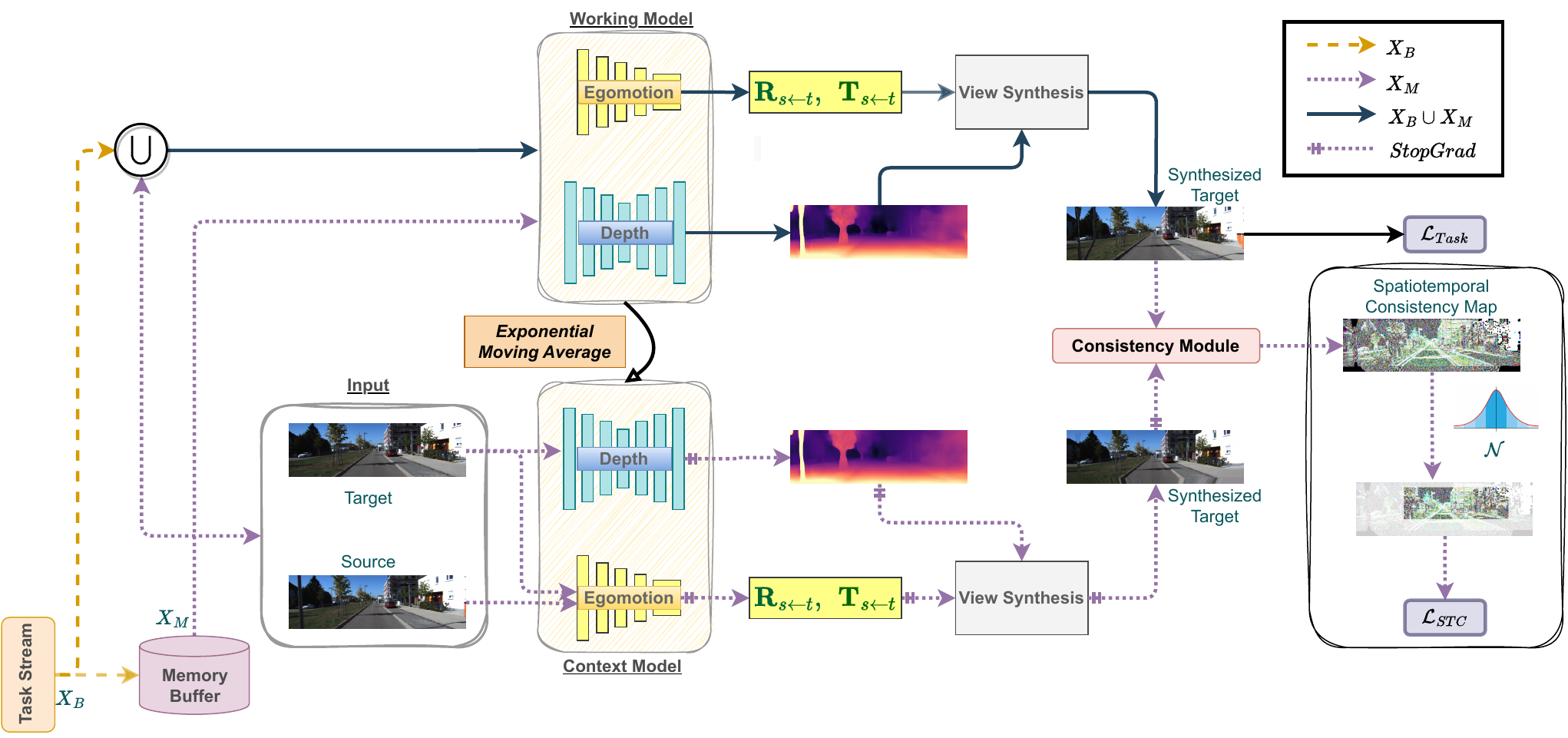}
\caption{Schematic of our MonoDepthCL: The context model consolidates the working model using an exponential moving average. The working model learns a view synthesis (See Section~\ref{sec:method_depth_estimation}) of the target image from the source images. A consistency module contrasts synthesized target images from both models to generate a spatiotemporal consistency map, which is cropped to a random ratio sampled from a Gaussian distribution to get a spatiotemporal consistency loss (See Algorithm~\ref{algo:crop_es}). 
Note that the depth estimation and associated operations are done at 4 resolutions.}
\label{fig:schematic}
\end{figure*}
 
Generally, multiple \textit{error} and \textit{accuracy} metrics are used to evaluate depth estimation performance~\cite{cadena2016measuring}. However, these metrics are not directly suitable for quantifying CL methods for depth estimation. Instead, by measuring the error and accuracy of depth estimation on each task after training a specific task, we generate a task-wise performance matrix $A \in \mathbb{R}^{n_t \times n_t}$, where $n_t$ is the total number of tasks. Consequently, we use the following metrics to evaluate CL models for depth estimation. Note that depending upon whether the metric is computed for the depth errors or the depth accuracy, the lower values or the higher values indicate better performance, respectively.

\textbf{Final Average} ($\mu_\text{final}$) measures the mean performance on all tasks after training the model on the final task. Hence, $\mu_\text{final} = \frac{1}{n_t}\sum_{j=1}^{n_t}A_{n_t, j}$. 

\textbf{Overall Average} ($\mu_\text{overall}$) measures the $\mu_\text{final}$ over all the seen tasks after training the model on each task. Hence, $\mu_\text{overall} = \frac{2}{n_t(n_t+1)}\sum_{i\geq j}^{n_t}A_{i, j}$. 
It indicates the improvements and degradations of model performance as the model is trained on different tasks. 

\textbf{Stability-Plasticity Trade-off} (SPTO) measures how well the model tackles the dilemma between retaining performance on the previously seen tasks and the capability to learn new tasks. Hence, SPTO $= \frac{2 \times A_\mathcal{S} \times A_\mathcal{P}}{A_\mathcal{S} + A_\mathcal{P}}$; where stability $A_\mathcal{S} = \frac{1}{n_t}\sum_{j=1}^{n_t-1}A_{n_t,j}$ 
is the average performance on all previously seen tasks after training the model on the final task, and plasticity $A_\mathcal{P}=\frac{1}{n_t}\sum_{i=1}^{n_t}A_{i,i}$ is the average performance of the tasks after the model is trained on them for the first time.

\section{Method}
\label{sec:method}

\begin{algorithm*}[t]
\caption{Algorithm for computing spatiotemporal consistency loss $\mathcal{L}_{\text{STC}}$.}
\label{algo:crop_es}
\small
\begin{algorithmic}[1]
\Statex {\bf Input:} {Per-pixel spatiotemporal consistencies from consistency module on $X_M$ for each source $j$ and prediction $i$, $STC^j_i \in \mathcal{R}^{H \times W} $}
\Statex {\bf Initialize: }{$\mathcal{L}_{\text{STC}}(X_M)=0.0$}
\For{$j \gets 1$ to $n_s$}
\State {$STC^j \gets 0.0$}
\For{$i \gets 1$ to $4$}
\State {$r \sim \mathcal{N}(0.5,0.1)$} \Comment{Sample ratio for cropping}
\State {$r \gets \texttt{Clip}(r, \texttt{min}=0.1, \texttt{max}=1.0)$} \Comment{Clip ratio between 0.1 and 1}
\State {$H^{'}, W^{'} \gets rH, rW$} \Comment{Height and width for cropping}
\State {$\text{\textit{top}} \gets \texttt{randint}(1, H^{'} - H + 1)$} \Comment{Sample start row for cropping}
\State {$\text{\textit{left}} \gets \texttt{randint}(1, W^{'} - W + 1)$} \Comment{Sample start column for cropping}
\State {$p^{'}_t \gets \{(x, y)\ \forall \ (x, y) \in [\text{\textit{top}}, \text{\textit{top}} + H^{'}-1] \times [\text{\textit{left}}, \text{\textit{left}} + W^{'} -1] \cap \mathcal{Z}\}$} \Comment{Set of cropped pixels}
\State {$STC^j \gets STC^j + \frac{1}{\lvert p^{'}_t\rvert}\sum_{p^{'}_t} STC^j_i[p^{'}_t]$} \Comment{Add average of cropped losses}
\EndFor
\State {$\mathcal{L}_{\text{STC}}(X_M) \gets \mathcal{L}_{\text{STC}}(X_M) + STC^j / 4$} \Comment{Add average across predictions}
\EndFor
\State {$\mathcal{L}_{\text{STC}}(X_M) \gets \mathcal{L}_{\text{STC}}(X_M) / n_s$} \Comment{Average across source images}
\Statex \Return{$\mathcal{L}_{\text{STC}}(X_M)$}
\end{algorithmic}
\end{algorithm*}
\normalsize

Here, we provide an overview of unsupervised monocular depth estimation, followed by our method for continual learning in depth estimation.
\label{sec:method_depth_estimation}

\subsection{Unsupervised Monocular Depth Estimation} 
Unsupervised monocular depth estimation is a technique used to determine the pixelwise distance of objects in a scene from a single unlabeled image. 
At any given training step, the input to the models is a set of temporally consecutive images, consisting of a \textit{target} image $I_t \in \mathbb{R}^{H\times W\times 3}$, and $n_s$ \textit{source} images $\{I^j_{s} \in \mathbb{R}^{H\times W\times 3}: j=1, 2,...n_s \}$, where $H$ and $W$ are the height and width, respectively, of the images. The depth network parameterized by $\theta_D$ predicts inverse depths at four resolutions, which are then bilinearly upsampled to the input resolution to reduce texture copy artifacts~\cite{godard2019digging}. Meanwhile, the ego-motion network parameterized by $\theta_E$ predicts the relative pose between each source-target image pair concatenated along the channel dimension.
Combining this relative pose with the camera intrinsics matrix, each source image $I^j_{s}$ is warped to the target image using the perspective projection equation~\cite{jaderberg2015spatial}.
This process, called \textit{view synthesis}, is performed for each upsampled target depth prediction indexed by $i=1,2,3,4$ to obtain a synthesized target image $\hat{I}^j_{i, t}$. An appearance-based photometric error is then formed between each synthesized target image and the original target image $I_{t}$. The photometric loss connects the predictions of the depth and ego-motion networks, forming the basis for unsupervised learning of depth. Additionally, to counteract the impact of temporally stationary pixels (e.g. when there is object motion but no ego-motion), the photometric loss is only considered at pixel locations where it is lower than the photometric loss between the unwarped source and target image at each scale. This procedure is known as automasking~\cite{godard2019digging}. Finally, a per-pixel edge-aware \textit{smoothness loss} is used to regularize the depth predictions~\cite{godard2017unsupervised}. The masked photometric loss and the smoothness loss together form the total training loss for unsupervised depth estimation, denoted by $\mathcal{L}_{depth}$ (see Supplementary Material for more details).

\subsection{Continual Learning for Unsupervised Monocular Depth Estimation}
\label{subsec:dualmem}

Humans continually learn from new experiences without catastrophically forgetting previous experiences~\cite{hadsell2020embracing}.
The complementary learning systems (CLS) theory postulates that human learning involves complex interactions between complementary learning systems that are learning at different rates~\cite{o2014complementary}. This includes a fast working system adapting quickly to new experiences and a slow context system consolidating knowledge from the fast systems. One such interaction could be the replay of sequences from memory such that the fast system works in the context of consolidated representations of the slow system~\cite{goyal2022inductive}. The fast and slow systems help model plasticity and stability, respectively.
Thus, we formulate a continual learning method for unsupervised monocular depth estimation with dual-memories and replay, which we call MonoDepthCL.

Concretely, consider \textit{working} depth and ego-motion model $\mathcal{WM}$ parameterized by $\theta^\mathcal{WM}=\theta^\mathcal{WM}_D \cup \theta^\mathcal{WM}_E$ and \textit{context} depth and ego-motion model $\mathcal{CM}$ parameterized by $\theta^\mathcal{CM}=\theta^\mathcal{CM}_D \cup \theta^\mathcal{CM}_E$. The working model is learnable, while the context model is maintained as an exponential moving average (EMA) of the working model~\cite{arani2022learning}. For replay, we employ a bounded \textit{memory} buffer $M$, updated by reservoir sampling~\cite{vitter1985random}, which allows the buffer to approximate the distribution of samples seen by the models~\cite{isele2018selective}. For an update coefficient $\alpha$ and update frequency $\nu \in (0, 1)$, we update the context models
in the training iteration $n$ to get,
\begin{equation} \label{eq:ema}
    \theta^\mathcal{CM} = \alpha_n \theta^\mathcal{CM} + (1 - \alpha_n) \theta^\mathcal{WM},
\end{equation}
where $\alpha_n = min(1 - 1 / (n + 1), \alpha)$.

However, this only helps with addressing forgetting in the context model. Ideally, the working model should have a mechanism to retain prior knowledge, which is learning in the context of consolidated representations~\cite{goyal2022inductive}. Furthermore, if the working model experiences catastrophic forgetting, it would also negatively affect the context model (Eq.~\ref{eq:ema}), which underscores the need for such a mechanism. Therefore, we distill the knowledge of consolidated representations of memory samples from the context model back to the working model. Since depth estimation involves training via view synthesis, we ensure consistency in the synthesized targets between the context and working models. This guarantees spatial consistency in the consolidated depth maps of the target images and temporal consistency in the poses between the target and nearby source images. We refer to this as spatiotemporal consistency.

\begin{figure*}[t]
\centering
\includegraphics[width=0.99\textwidth]{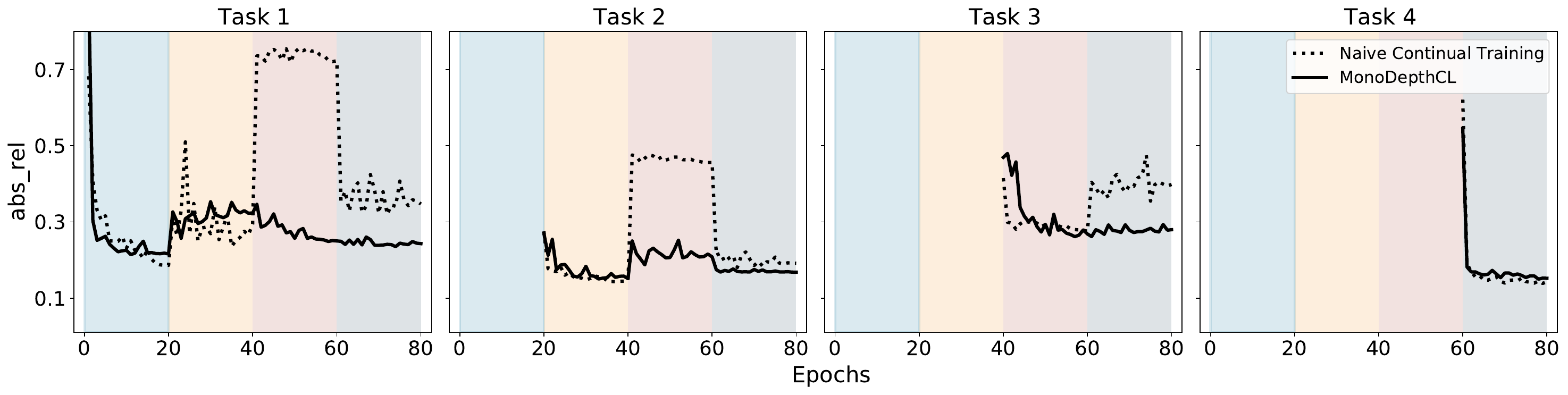}
\caption{Performance on the tasks over epochs as models are trained with buffer size 200. Naive Continual training undergoes catastrophic forgetting, while MonoDepthCL mitigates it to a good extent.} 
\label{fig:training_graphs}
\end{figure*}

Specifically, at each training step, we sample a batch $X_B$ from the current task stream and a batch $X_M$ from the memory buffer. On each memory sample, we warp the source images to the target image 
using both the working and context models. Let $\hat{I}^{j, \mathcal{CM}}_{i,t}$ and $\hat{I}^{j, \mathcal{WM}}_{i, t}$ represent the synthesized targets for the $j^{\text{th}}$ source image and the $i^{\text{th}}$ depth prediction using context and working models, respectively. Then, a consistency module computes a per-pixel spatiotemporal consistency between these synthesized targets as follows:
\begin{equation}
\label{eq:stc_terms}
\begin{split}
STC^j_i =& \frac{\rho}{2}[1 - SSIM(\hat{I}^{j, \mathcal{CM}}_{i, t}, \hat{I}^{j, \mathcal{WM}}_{i, t})] \\ 
&+ (1 - \rho)\left| \hat{I}^{j, \mathcal{CM}}_{i, t} - \hat{I}^{j, \mathcal{WM}}_{i, t} \right|,
\end{split}
\end{equation}
where SSIM refers to the Structural Similarity Index between two images~\cite{wang2004image}.
Each per-pixel consistency $STC^j_i$ is an image-shaped loss of shape $H$$\times$$W$. To reduce the overfitting to buffer samples~\cite{buzzega2021rethinking}, improve invariance through augmentation~\cite{purushwalkam2020demystifying}, and improve efficiency, we randomly crop these per-pixel consistencies before evaluating the final spatiotemporal consistency loss $\mathcal{L}_{\text{STC}}$. We sample the ratios to be cropped for each consistency from a Gaussian distribution with $0.5$ mean and $0.1$ standard deviation. This would retain half the $H$$\times$$W$ consistency term on average, and retain 20-80\% of the term $\sim$99.7\% of the times (i.e. within $3$ standard deviations). These cropped losses are then averaged across all predictions, source images, and cropped pixels to get $\mathcal{L}_{\text{STC}}$.
Algorithm~\ref{algo:crop_es} shows the process to compute the cropped spatiotemporal consistency loss. 
Note that we use a warmup for the spatiotemporal consistency loss, such that it is only applied after the first task is learned. This allows the view synthesis to be learned well before it is used as a constraint between the context and working models.

Finally, the working model needs to learn the new task and is thus trained with a task loss $\mathcal{L}_{Task}$, which is the depth loss $\mathcal{L}_{depth}$ 
discussed earlier,
on the union of current and memory batches. Putting everything together, the total loss for continual unsupervised monocular depth estimation is:
\begin{equation}
    \label{eq:total_general_loss}
    \mathcal{L}_{\text{total}} = \mathcal{L}_{Task}(X_B \cup X_M) + \beta \mathcal{L}_{\text{STC}}(X_M). 
\end{equation}
We dub our method as \textit{\textbf{Mono}cular \textbf{Depth} estimation with \textbf{C}ontinual \textbf{L}earning} or \textit{\textbf{MonoDepthCL}}. The complete schematic of our method can be seen in Figure~\ref{fig:schematic}.



\begin{table*}[t]
\centering
\small
\begin{tabular}{c|c|ccc|ccc|ccc}
\hline
\multirow{2}{*}{\textbf{Buffer}} & \multirow{2}{*}{\textbf{Method}} & \multicolumn{3}{c|}{$\mathbf{\mu_\textbf{final}}$} & \multicolumn{3}{c|}{$\mathbf{\mu_\textbf{overall}}$} & \multicolumn{3}{c}{\textbf{SPTO}} 
\\ \cline{3-11} 
 & & {\cellcolor{red!20}abs\_rel$\downarrow$} & {\cellcolor{red!20}RMSE$\downarrow$} & {\cellcolor{blue!20}a1$\uparrow$} & {\cellcolor{red!20}abs\_rel$\downarrow$}  & {\cellcolor{red!20}RMSE$\downarrow$}  & {\cellcolor{blue!20}a1$\uparrow$}  & {\cellcolor{red!20}abs\_rel$\downarrow$} & {\cellcolor{red!20}RMSE$\downarrow$} & {\cellcolor{blue!20}a1$\uparrow$} 
 \\ \hline \hline
\multirow{2}{*}{--}& Joint & 0.177  & 5.408  & 0.759  & \multicolumn{3}{c|}{--} & \multicolumn{3}{c}{--} \\ 
& NCT & 0.272  & 7.397  & 0.639  & 0.315  & 8.253  & 0.580  & 0.238  & 6.313  & 0.648 \\
\hline \hline
\multirow{3}{*}{50} & ER  & 0.289  & 7.421  & 0.619  & 0.310  & 7.770  & 0.580  & 0.266  & 6.422  & 0.625 \\
& ContextDepth & 0.274  & 7.310  & 0.631  & 0.290  & 7.669  & 0.594  & 0.242 & 6.287  & 0.644 \\
& MonoDepthCL & \textbf{0.249} & \textbf{6.774} & \textbf{0.647} & \textbf{0.265} & \textbf{7.168} & \textbf{0.618} & \textbf{0.237} & \textbf{5.987} & \textbf{0.652} \\ \hline
\multirow{3}{*}{200} & ER & 0.248 & 6.995 & 0.666 & 0.238 & 6.852 & \textbf{0.679} & 0.228 & 6.047 & 0.673 \\
& ContextDepth & 0.286 & 7.177 & 0.644 & 0.296 & 7.368 & 0.621 & 0.265 & 6.271 & 0.644 \\
& MonoDepthCL & \textbf{0.228} & \textbf{6.583} & \textbf{0.673} & \textbf{0.237} & \textbf{6.753} & 0.663 & \textbf{0.223} & \textbf{5.832} & \textbf{0.676} \\
\hline
\multirow{3}{*}{500} & ER & 0.261 & 6.879 & 0.680 & 0.236 & 6.586 & 0.691 & 0.242 & 6.033 & 0.678 \\
& ContextDepth & 0.252 & 6.695 & 0.677 & 0.239 & 6.585 & 0.677 & 0.242 & 5.932 & 0.672 \\
& MonoDepthCL & \textbf{0.228} & \textbf{6.278} & \textbf{0.693} & \textbf{0.219} & \textbf{6.239} & \textbf{0.701} & \textbf{0.222} & \textbf{5.573} & \textbf{0.693} \\ \hline 
\end{tabular}
\caption{Performance of different methods on the CUDE framework for multiple buffer sizes. The best results for each buffer size are shown in bold.}
\label{tab:ablation}
\end{table*}

\section{Results}
\label{sec:experiments}

We demonstrate CL challenges in unsupervised monocular depth estimation on the CUDE framework and the effectiveness of MonoDepthCL for mitigating catastrophic forgetting. The architecture details and hyperparameters used can be found in the
Supplementary Material.
We compare against naive continual training (NCT), and training on the whole dataset, i.e. its joint distribution after collecting data from all tasks, dubbed \textit{Joint}, following standard practice in the CL literature \cite{buzzega2020dark,DBLP:conf/iclr/RiemerCALRTT19}. NCT and Joint form the lower and upper bounds for CL, respectively. Our method has two models connected by a spatiotemporal consistency loss. Consequently, we consider two additional CL methods stemming from these models in the absence of the proposed spatiotemporal consistency loss. The fast-learning working model, if trained in isolation, would learn merely on the task loss with rehearsal, and should be capable of mitigating forgetting to some extent. This is equivalent to ER (which has already been shown to outperform most other rehearsal-based methods for classification~\cite{van2022three}). Similarly, the context model, which maintains an exponential moving average of the working model, should also be capable of mitigating forgetting to some extent. We call this method ContextDepth.

Figure~\ref{fig:training_graphs} shows a significant drop in the performance of NCT on each task as newer tasks are encountered, undergoing catastrophic forgetting as seen earlier in Figure~\ref{fig:problem}. This is in contrast to MonoDepthCL, which improves over NCT, undergoing far less forgetting across tasks.
Table~\ref{tab:ablation} additionally shows that MonoDepthCL outperforms other CL methods in all buffer sizes across all metrics, demonstrating the effectiveness of the spatiotemporal consistency loss.

\begin{figure}[t]
\centering
\includegraphics[width=0.48\textwidth]{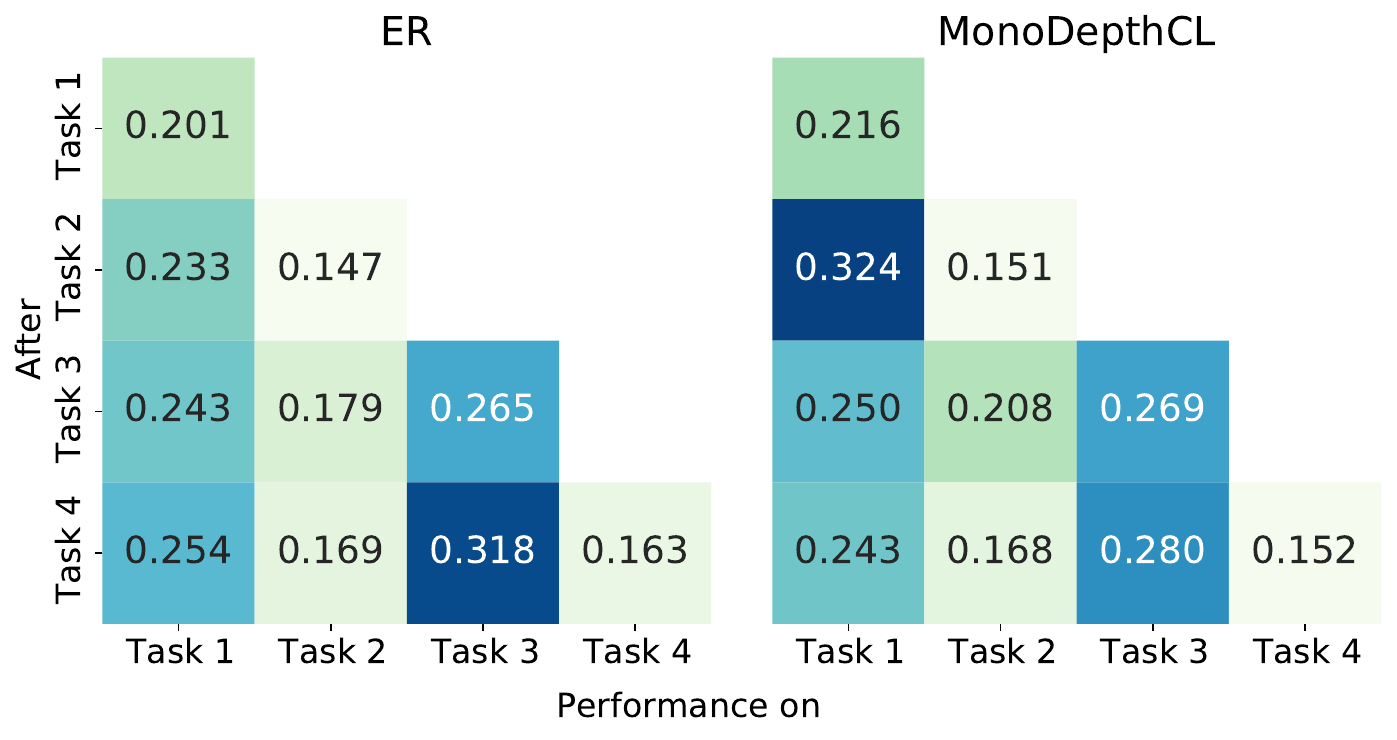}
\caption{Taskwise errors after training on each task for buffer size 200. A more uniform distribution of color in a row indicates a lower task-task recency bias after learning the task corresponding to that row. ER shows a higher bias towards recent tasks.}
\label{fig:taskwise_heatmaps}
\end{figure}

Increasing the buffer size leads to a general improvement across metrics for all CL methods, including MonoDepthCL. However, at low buffer size ($50$), ER and ContextDepth fall behind even NCT on some SPTO and $\mathbf{\mu_\textbf{final}}$ metrics. Nevertheless, they outperform NCT on $\mathbf{\mu_\textbf{overall}}$. This is because $\mathbf{\mu_\textbf{overall}}$ measures the improvements and degradations of model performance across the task trajectory, and not just the mean accuracy after learning the final task. 
When rehearsal and dual model approach are combined with our spatiotemporal consistency loss, MonoDepthCL performs well on \textit{all} metrics for the low buffer size as well. 
Consequently, we hypothesize that the higher performance of MonoDepthCL on the $\mathbf{\mu_\textbf{overall}}$ metrics indicates its ability to learn on additional tasks.
The performance of MonoDepthCL on the SPTO metrics further confirms that it is better equipped to handle the stability-plasticity trade-off compared to rehearsal-based ER. The stronger SPTO performance of MonoDepthCL also translates to lower task-recency bias as seen in Figure~\ref{fig:taskwise_heatmaps}. The distribution of performance across previous tasks after learning each task is more uniform for our method than for ER. This is in line with recent findings in image classification~\cite{arani2022learning}.

Hence, benchmarking the methods on our CUDE framework demonstrates the challenge of CL for unsupervised monocular depth estimation. 
Our experiments additionally show that the metrics defined in CUDE capture different aspects of CL. 
We contend that our method with its spatiotemporal consistency loss is an effective strategy for handling the stability-plasticity trade-off in CL.

\begin{table}[t]
\centering
\small
\begin{tabular}{c|ccc}
\hline
\multirow{2}{*}{\textbf{Method}} & \multicolumn{3}{c}{$\mathbf{\mu_\textbf{final}}$}
\\ \cline{2-4}
 & {\cellcolor{red!20}abs\_rel$\downarrow$} & {\cellcolor{red!20}RMSE$\downarrow$} & {\cellcolor{blue!20}a1$\uparrow$} 
 \\ \hline \hline
Joint & 0.180  & 5.455 & 0.752  \\ 
NCT  & 0.264  & 7.280  & 0.640 \\ \hline 
MonoDepthCL & 0.252  & 7.237  & 0.663 \\ \hline 
\end{tabular}
\caption{Camera intrinsics $K$ are learned during training with a buffer size 200. MonoDepthCL reduces catastrophic forgetting even without prior knowledge of camera intrinsics.}
\label{tab:learnt}
\end{table}

\begin{table*}[t]
\centering
\small
\begin{tabular}{c|c|ccc|ccc|ccc}
\hline
\multirow{2}{*}{\textbf{Random cropping}} & \multirow{2}{*}{\textbf{Warmup}} & \multicolumn{3}{c|}{$\mathbf{\mu_\textbf{final}}$} & \multicolumn{3}{c|}{$\mathbf{\mu_\textbf{overall}}$} & \multicolumn{3}{c}{\textbf{SPTO}} 
\\ \cline{3-11} 
 & & {\cellcolor{red!20}abs\_rel$\downarrow$} & {\cellcolor{red!20}RMSE$\downarrow$} & {\cellcolor{blue!20}a1$\uparrow$} & {\cellcolor{red!20}abs\_rel$\downarrow$}  & {\cellcolor{red!20}RMSE$\downarrow$}  & {\cellcolor{blue!20}a1$\uparrow$}  & {\cellcolor{red!20}abs\_rel$\downarrow$} & {\cellcolor{red!20}RMSE$\downarrow$} & {\cellcolor{blue!20}a1$\uparrow$} 
 \\ \hline \hline
 \xmark & \xmark & 0.306  & 7.679  & 0.625  & 0.333  & 8.187  & 0.597  & 0.290  & 6.837  & 0.621 \\
 \xmark & \cmark & 0.237  & 6.623  & 0.669  & 0.239  & \textbf{6.707}  & \textbf{0.663}  & 0.231  & 5.869  & 0.669 \\
 \cmark & \cmark & \textbf{0.228}  & \textbf{6.583}  & \textbf{0.673}  & \textbf{0.237}  & 6.753  & \textbf{0.663}  & \textbf{0.223}  & \textbf{5.832}  & \textbf{0.676} \\
 \hline
\end{tabular}
\caption{Ablation of design components of the spatiotemporal consistency loss on the CUDE framework for buffer size 200. The best results are shown in bold. }
\label{tab:stc_ablation}
\end{table*}

\begin{table*}[t]
\centering
\small
\begin{tabular}{c|ccc|ccc|ccc}
\hline
\multirow{2}{*}{\textbf{Method}} & \multicolumn{3}{c|}{$\mathbf{\mu_\textbf{final}}$} & \multicolumn{3}{c|}{$\mathbf{\mu_\textbf{overall}}$} & \multicolumn{3}{c}{\textbf{SPTO}} 
\\ \cline{2-10}
 & {\cellcolor{red!20}abs\_rel$\downarrow$} & {\cellcolor{red!20}RMSE$\downarrow$} & {\cellcolor{blue!20}a1$\uparrow$} & {\cellcolor{red!20}abs\_rel$\downarrow$}  & {\cellcolor{red!20}RMSE$\downarrow$}  & {\cellcolor{blue!20}a1$\uparrow$}  & {\cellcolor{red!20}abs\_rel$\downarrow$} & {\cellcolor{red!20}RMSE$\downarrow$} & {\cellcolor{blue!20}a1$\uparrow$} \\ \hline \hline
Joint & 0.182 & 7.413 & 0.755 & \multicolumn{3}{c|}{--} & \multicolumn{3}{c}{--}\\ 
NCT & 0.316  & 9.966 & 0.631 & 0.317 & 8.788 & 0.597 & 0.255 & 7.571 & 0.64 \\
\hline \hline
ER & 0.289  & 9.618 & 0.637 & 0.264 & 7.782 & 0.649 & 0.255 & 7.388 & 0.637 \\
ContextDepth & 0.293  & 9.666 & 0.635 & 0.282 & 8.026 & 0.631 & 0.264 & 7.439 & 0.631 \\
MonoDepthCL & \textbf{0.246}  & \textbf{8.626} & \textbf{0.664} & \textbf{0.242} & \textbf{7.353} & \textbf{0.656} & \textbf{0.237} & \textbf{6.773} & \textbf{0.655} 
\\ \hline
\end{tabular}
\caption{Performance of different methods on the CUDE-5 framework for buffer-size 200. The best results are shown in bold.}
\label{tab:appendix_ablation}
\end{table*}

\textbf{With Learned Intrinsics:} Now, unsupervised monocular depth estimation requires knowledge of the camera intrinsics for the perspective projection (Section~\ref{sec:method_depth_estimation}). 
This forms a roadblock for training on crowdsourced video sequences where this information is not available. While recent research has demonstrated unsupervised monocular depth estimation with learned camera intrinsics in joint training, it has not been explored in the CL setting. 
We train our proposed MonoDepthCL with learned intrinsics and benchmark it on the CUDE framework. The results in Table~\ref{tab:learnt} demonstrate that MonoDepthCL outperforms NCT across all $\mathbf{\mu_\textbf{final}}$ metrics. Although a considerable gap still exists between Joint and MonoDepthCL, our study represents the pioneering effort towards crowdsourced CL for unsupervised monocular depth estimation.

\subsection{Ablation Study}
\label{sec:ablation}

Our spatiotemporal consistency loss has two major design components - warmup during the first task to allow the view synthesis to be learned properly before it is applied as a constraint between the working and context models; and random cropping of the spatiotemporal consistency map to introduce invariance through augmentation and improve efficiency.
Table~\ref{tab:stc_ablation} details the impact of both of these design components on the continual learning metrics. It can be seen that warmup contributes to a great improvement in continual learning performance, and adding random cropping leads to a further improvement over warmup. 
Therefore, both the design components of our spatiotemporal consistency loss have a significant impact on the continual learning performance.

\subsection{Longer Task Sequence (CUDE-5)}
\label{appendix_empirical_cude5}

In Section~\ref{sec:experiments}, we noted that MonoDepthCL is expected to perform well on longer task sequences as well. Accordingly, we extend the framework CUDE to 5 tasks with an additional task from dataset DDAD~\cite{guizilini20203d} as shown in the Supplementary Material (Figure ~\ref{fig:framework-5}). 
Since DDAD contains videos from USA and Japan, and captures the domain shifts from one country to another. Additionally, the ground truth depth range for DDAD is 200m which is double than that of Cityscapes. 
We report the performance on CUDE-5 for buffer size 200 in Table~\ref{tab:appendix_ablation}. We observe that MonoDepthCL continues to mitigate forgetting and improve performance through spatiotemporal consistency.

\textbf{With Learned Intrinsics:} Similar to CUDE with 4 tasks, we also evaluate the performance of MonoDepthCL when the camera intrinsics of all 5 tasks are learned together with the depth in Table~\ref{tab:appendix_learned}. 
The ability to continually learn depth estimation from additional data, without prior knowledge of camera intrinsics paves the way towards crowdsourced depth estimation. 

\section{Conclusion}
\label{sec:conclusion}

We introduce CUDE, a framework for benchmarking CL methods for unsupervised monocular depth estimation, along with our MonoDepthCL method.
CUDE consists of four sequential tasks that span different weather and lighting conditions, depth ranges, and navigation scenarios for indoor and outdoor scenes. It also defines metrics that measure the final average performance after learning the final task, the overall average performance throughout the learning trajectory, and the stability-plasticity trade-off. 
Meanwhile, our proposed method follows a dual-model approach with a memory buffer for storing previously seen information. Specifically, a working model learns the task on samples from memory and data stream, while a context model is maintained as an exponential moving average of the working model. A novel spatiotemporal consistency loss efficiently enforces the view synthesis consistency between the two models. 
The benchmarking of MonoDepthCL on CUDE shows the effectiveness of our method and the spatiotemporal consistency loss for CL, and in dealing with the stability-plasticity trade-off. It also demonstrates the value of the framework itself, including the defined metrics that capture various aspects of CL performance. 
Finally, we show the applicability of MonoDepthCL for a scenario where the camera intrinsics also need to be learned.

We put forward our CUDE framework as a first step towards comprehensive benchmarking of CL methods for unsupervised monocular depth estimation. We further contend that our method, MonoDepthCL provides a promising approach towards addressing the performance gap between continual training and joint training.

\begin{table}[t]
\centering
\small
\begin{tabular}{c|ccc}
\hline
\multirow{2}{*}{\textbf{Method}} & \multicolumn{3}{c}{$\mathbf{\mu_\textbf{final}}$} \\ \cline{2-4}
 & {\cellcolor{red!20}abs\_rel$\downarrow$} & {\cellcolor{red!20}RMSE$\downarrow$} & {\cellcolor{blue!20}a1$\uparrow$} \\ \hline \hline
Joint & 0.185  & 7.699 & 0.751  \\ 
NCT  & 0.315  & 9.868  & 0.614  \\ \hline 
MonoDepthCL & 0.252  & 7.237  & 0.663 \\ \hline 
\end{tabular}
\caption{$\mu_\text{final}$ performance on the CUDE-5 framework when the camera intrinsics K are learned during training. MonoDepthCL reduces catastrophic forgetting even without prior knowledge of camera intrinsics on the longer task sequence as well.}
\label{tab:appendix_learned}
\end{table}

\paragraph{\textbf{Acknowledgement:}}
The work was conducted while all the authors were affiliated with NavInfo Europe, Eindhoven, The Netherlands.
{\small
\bibliographystyle{ieee_fullname}
\bibliography{WACV/ref}

\begin{thebibliography}{10}\itemsep=-1pt

\bibitem{abraham2005memory}
Wickliffe~C Abraham and Anthony Robins.
\newblock Memory retention--the synaptic stability versus plasticity dilemma.
\newblock {\em Trends in neurosciences}, 28(2):73--78, 2005.

\bibitem{adel2019continual}
Tameem Adel, Han Zhao, and Richard~E Turner.
\newblock Continual learning with adaptive weights (claw).
\newblock {\em arXiv preprint arXiv:1911.09514}, 2019.

\bibitem{arani2022learning}
Elahe Arani, Fahad Sarfraz, and Bahram Zonooz.
\newblock Learning fast, learning slow: A general continual learning method
  based on complementary learning system.
\newblock In {\em International Conference on Learning Representations}, 2022.

\bibitem{bian2019unsupervised}
Jiawang Bian, Zhichao Li, Naiyan Wang, Huangying Zhan, Chunhua Shen, Ming-Ming
  Cheng, and Ian Reid.
\newblock Unsupervised scale-consistent depth and ego-motion learning from
  monocular video.
\newblock {\em Advances in neural information processing systems}, 32, 2019.

\bibitem{bian2021auto}
Jia-Wang Bian, Huangying Zhan, Naiyan Wang, Tat-Jun Chin, Chunhua Shen, and Ian
  Reid.
\newblock Auto-rectify network for unsupervised indoor depth estimation.
\newblock {\em IEEE Transactions on Pattern Analysis and Machine Intelligence},
  44(12):9802--9813, 2021.

\bibitem{boschini2022class}
Matteo Boschini, Lorenzo Bonicelli, Pietro Buzzega, Angelo Porrello, and Simone
  Calderara.
\newblock Class-incremental continual learning into the extended der-verse.
\newblock {\em IEEE Transactions on Pattern Analysis and Machine Intelligence},
  2022.

\bibitem{butler2015privacy}
Daniel~J Butler, Justin Huang, Franziska Roesner, and Maya Cakmak.
\newblock The privacy-utility tradeoff for remotely teleoperated robots.
\newblock In {\em Proceedings of the tenth annual ACM/IEEE international
  conference on human-robot interaction}, pages 27--34, 2015.

\bibitem{buzzega2020dark}
Pietro Buzzega, Matteo Boschini, Angelo Porrello, Davide Abati, and Simone
  Calderara.
\newblock Dark experience for general continual learning: a strong, simple
  baseline.
\newblock {\em Advances in neural information processing systems},
  33:15920--15930, 2020.

\bibitem{buzzega2021rethinking}
Pietro Buzzega, Matteo Boschini, Angelo Porrello, and Simone Calderara.
\newblock Rethinking experience replay: a bag of tricks for continual learning.
\newblock In {\em 2020 25th International Conference on Pattern Recognition
  (ICPR)}, pages 2180--2187. IEEE, 2021.

\bibitem{cabon2020vkitti2}
Yohann Cabon, Naila Murray, and Martin Humenberger.
\newblock Virtual kitti 2, 2020.

\bibitem{cadena2016measuring}
Cesar Cadena, Yasir Latif, and Ian~D Reid.
\newblock Measuring the performance of single image depth estimation methods.
\newblock In {\em 2016 IEEE/RSJ International Conference on Intelligent Robots
  and Systems (IROS)}, pages 4150--4157. IEEE, 2016.

\bibitem{chawla2021multimodal}
Hemang Chawla, Arnav Varma, Elahe Arani, and Bahram Zonooz.
\newblock Multimodal scale consistency and awareness for monocular
  self-supervised depth estimation.
\newblock In {\em 2021 IEEE International Conference on Robotics and Automation
  (ICRA)}, pages 5140--5146. IEEE, 2021.

\bibitem{cordts2016cityscapes}
Marius Cordts, Mohamed Omran, Sebastian Ramos, Timo Rehfeld, Markus Enzweiler,
  Rodrigo Benenson, Uwe Franke, Stefan Roth, and Bernt Schiele.
\newblock The cityscapes dataset for semantic urban scene understanding.
\newblock In {\em Proceedings of the IEEE conference on computer vision and
  pattern recognition}, pages 3213--3223, 2016.

\bibitem{douillard2021plop}
Arthur Douillard, Yifu Chen, Arnaud Dapogny, and Matthieu Cord.
\newblock Plop: Learning without forgetting for continual semantic
  segmentation.
\newblock In {\em Proceedings of the IEEE/CVF Conference on Computer Vision and
  Pattern Recognition}, pages 4040--4050, 2021.

\bibitem{eigen2014depth}
David Eigen, Christian Puhrsch, and Rob Fergus.
\newblock Depth map prediction from a single image using a multi-scale deep
  network.
\newblock {\em Advances in neural information processing systems}, 27, 2014.

\bibitem{farquhar_towards_2018}
Sebastian Farquhar and Yarin Gal.
\newblock Towards {Robust} {Evaluations} of {Continual} {Learning}.
\newblock {\em Lifelong Learning: A Reinforcement Learning Approach Workshop at
  ICML}, 2018.

\bibitem{geiger2013vision}
Andreas Geiger, Philip Lenz, Christoph Stiller, and Raquel Urtasun.
\newblock Vision meets robotics: The kitti dataset.
\newblock {\em The International Journal of Robotics Research},
  32(11):1231--1237, 2013.

\bibitem{godard2017unsupervised}
Cl{\'e}ment Godard, Oisin Mac~Aodha, and Gabriel~J Brostow.
\newblock Unsupervised monocular depth estimation with left-right consistency.
\newblock In {\em Proceedings of the IEEE conference on computer vision and
  pattern recognition}, pages 270--279, 2017.

\bibitem{godard2019digging}
Cl{\'e}ment Godard, Oisin Mac~Aodha, Michael Firman, and Gabriel~J Brostow.
\newblock Digging into self-supervised monocular depth estimation.
\newblock In {\em Proceedings of the IEEE/CVF International Conference on
  Computer Vision}, pages 3828--3838, 2019.

\bibitem{gordon2019depth}
Ariel Gordon, Hanhan Li, Rico Jonschkowski, and Anelia Angelova.
\newblock Depth from videos in the wild: Unsupervised monocular depth learning
  from unknown cameras.
\newblock In {\em Proceedings of the IEEE/CVF International Conference on
  Computer Vision}, pages 8977--8986, 2019.

\bibitem{goyal2022inductive}
Anirudh Goyal and Yoshua Bengio.
\newblock Inductive biases for deep learning of higher-level cognition.
\newblock {\em Proceedings of the Royal Society A}, 478(2266):20210068, 2022.

\bibitem{guizilini2021sparse}
Vitor Guizilini, Rares Ambrus, Wolfram Burgard, and Adrien Gaidon.
\newblock Sparse auxiliary networks for unified monocular depth prediction and
  completion.
\newblock In {\em Proceedings of the IEEE/CVF Conference on Computer Vision and
  Pattern Recognition}, pages 11078--11088, 2021.

\bibitem{guizilini20203d}
Vitor Guizilini, Rares Ambrus, Sudeep Pillai, Allan Raventos, and Adrien
  Gaidon.
\newblock 3d packing for self-supervised monocular depth estimation.
\newblock In {\em Proceedings of the IEEE/CVF conference on computer vision and
  pattern recognition}, pages 2485--2494, 2020.

\bibitem{guizilini2022multi}
Vitor Guizilini, Rareș Ambruș, Dian Chen, Sergey Zakharov, and Adrien Gaidon.
\newblock Multi-frame self-supervised depth with transformers.
\newblock In {\em Proceedings of the IEEE/CVF Conference on Computer Vision and
  Pattern Recognition}, pages 160--170, 2022.

\bibitem{hadsell2020embracing}
Raia Hadsell, Dushyant Rao, Andrei~A Rusu, and Razvan Pascanu.
\newblock Embracing change: Continual learning in deep neural networks.
\newblock {\em Trends in cognitive sciences}, 24(12):1028--1040, 2020.

\bibitem{hu2020probabilistic}
Anthony Hu, Fergal Cotter, Nikhil Mohan, Corina Gurau, and Alex Kendall.
\newblock Probabilistic future prediction for video scene understanding.
\newblock In {\em Computer Vision--ECCV 2020: 16th European Conference,
  Glasgow, UK, August 23--28, 2020, Proceedings, Part XVI 16}, pages 767--785.
  Springer, 2020.

\bibitem{isele2018selective}
David Isele and Akansel Cosgun.
\newblock Selective experience replay for lifelong learning.
\newblock In {\em Proceedings of the AAAI Conference on Artificial
  Intelligence}, 2018.

\bibitem{jaderberg2015spatial}
Max Jaderberg, Karen Simonyan, Andrew Zisserman, et~al.
\newblock Spatial transformer networks.
\newblock {\em Advances in neural information processing systems}, 28, 2015.

\bibitem{kalia2019real}
Megha Kalia, Nassir Navab, and Tim Salcudean.
\newblock A real-time interactive augmented reality depth estimation technique
  for surgical robotics.
\newblock In {\em 2019 International Conference on Robotics and Automation
  (ICRA)}, pages 8291--8297. IEEE, 2019.

\bibitem{kang2022forget}
Haeyong Kang, Rusty John~Lloyd Mina, Sultan Rizky~Hikmawan Madjid, Jaehong
  Yoon, Mark Hasegawa-Johnson, Sung~Ju Hwang, and Chang~D Yoo.
\newblock Forget-free continual learning with winning subnetworks.
\newblock In {\em International Conference on Machine Learning}, pages
  10734--10750. PMLR, 2022.

\bibitem{kirkpatrick2017overcoming}
James Kirkpatrick, Razvan Pascanu, Neil Rabinowitz, Joel Veness, Guillaume
  Desjardins, Andrei~A Rusu, Kieran Milan, John Quan, Tiago Ramalho, Agnieszka
  Grabska-Barwinska, et~al.
\newblock Overcoming catastrophic forgetting in neural networks.
\newblock {\em Proceedings of the national academy of sciences},
  114(13):3521--3526, 2017.

\bibitem{kuznietsov2021comoda}
Yevhen Kuznietsov, Marc Proesmans, and Luc Van~Gool.
\newblock Comoda: Continuous monocular depth adaptation using past experiences.
\newblock In {\em Proceedings of the IEEE/CVF Winter Conference on Applications
  of Computer Vision}, pages 2907--2917, 2021.

\bibitem{ladicky2014pulling}
Lubor Ladicky, Jianbo Shi, and Marc Pollefeys.
\newblock Pulling things out of perspective.
\newblock In {\em Proceedings of the IEEE conference on computer vision and
  pattern recognition}, pages 89--96, 2014.

\bibitem{lee2021patch}
Sihaeng Lee, Janghyeon Lee, Byungju Kim, Eojindl Yi, and Junmo Kim.
\newblock Patch-wise attention network for monocular depth estimation.
\newblock In {\em Proceedings of the AAAI Conference on Artificial
  Intelligence}, volume~35, pages 1873--1881, 2021.

\bibitem{lesort2020continual}
Timoth{\'e}e Lesort, Vincenzo Lomonaco, Andrei Stoian, Davide Maltoni, David
  Filliat, and Natalia D{\'\i}az-Rodr{\'\i}guez.
\newblock Continual learning for robotics: Definition, framework, learning
  strategies, opportunities and challenges.
\newblock {\em Information fusion}, 58:52--68, 2020.

\bibitem{lyu2021hr}
Xiaoyang Lyu, Liang Liu, Mengmeng Wang, Xin Kong, Lina Liu, Yong Liu, Xinxin
  Chen, and Yi Yuan.
\newblock Hr-depth: High resolution self-supervised monocular depth estimation.
\newblock In {\em Proceedings of the AAAI Conference on Artificial
  Intelligence}, 2021.

\bibitem{maracani2021recall}
Andrea Maracani, Umberto Michieli, Marco Toldo, and Pietro Zanuttigh.
\newblock Recall: Replay-based continual learning in semantic segmentation.
\newblock In {\em Proceedings of the IEEE/CVF International Conference on
  Computer Vision}, pages 7026--7035, 2021.

\bibitem{mccloskey1989catastrophic}
Michael McCloskey and Neal~J Cohen.
\newblock Catastrophic interference in connectionist networks: The sequential
  learning problem.
\newblock In {\em Psychology of learning and motivation}, volume~24, pages
  109--165. Elsevier, 1989.

\bibitem{ming2021deep}
Yue Ming, Xuyang Meng, Chunxiao Fan, and Hui Yu.
\newblock Deep learning for monocular depth estimation: A review.
\newblock {\em Neurocomputing}, 438:14--33, 2021.

\bibitem{o2014complementary}
Randall~C O’Reilly, Rajan Bhattacharyya, Michael~D Howard, and Nicholas Ketz.
\newblock Complementary learning systems.
\newblock {\em Cognitive science}, 38(6):1229--1248, 2014.

\bibitem{patil2022p3depth}
Vaishakh Patil, Christos Sakaridis, Alexander Liniger, and Luc Van~Gool.
\newblock P3depth: Monocular depth estimation with a piecewise planarity prior.
\newblock In {\em Proceedings of the IEEE/CVF Conference on Computer Vision and
  Pattern Recognition}, pages 1610--1621, 2022.

\bibitem{pham2021dualnet}
Quang Pham, Chenghao Liu, and Steven Hoi.
\newblock Dualnet: Continual learning, fast and slow.
\newblock {\em Advances in Neural Information Processing Systems},
  34:16131--16144, 2021.

\bibitem{purushwalkam2020demystifying}
Senthil Purushwalkam and Abhinav Gupta.
\newblock Demystifying contrastive self-supervised learning: Invariances,
  augmentations and dataset biases.
\newblock {\em Advances in Neural Information Processing Systems},
  33:3407--3418, 2020.

\bibitem{ranftl2021vision}
Ren{\'e} Ranftl, Alexey Bochkovskiy, and Vladlen Koltun.
\newblock Vision transformers for dense prediction.
\newblock In {\em Proceedings of the IEEE/CVF International Conference on
  Computer Vision}, pages 12179--12188, 2021.

\bibitem{DBLP:conf/iclr/RiemerCALRTT19}
Matthew Riemer, Ignacio Cases, Robert Ajemian, Miao Liu, Irina Rish, Yuhai Tu,
  and Gerald Tesauro.
\newblock Learning to learn without forgetting by maximizing transfer and
  minimizing interference.
\newblock In {\em 7th International Conference on Learning Representations,
  {ICLR} 2019, New Orleans, LA, USA, May 6-9, 2019}, 2019.

\bibitem{rusu2017sim}
Andrei~A Rusu, Matej Ve{\v{c}}er{\'\i}k, Thomas Roth{\"o}rl, Nicolas Heess,
  Razvan Pascanu, and Raia Hadsell.
\newblock Sim-to-real robot learning from pixels with progressive nets.
\newblock In {\em Conference on robot learning}, pages 262--270. PMLR, 2017.

\bibitem{schonberger2016structure}
Johannes~L Schonberger and Jan-Michael Frahm.
\newblock Structure-from-motion revisited.
\newblock In {\em Proceedings of the IEEE conference on computer vision and
  pattern recognition}, pages 4104--4113, 2016.

\bibitem{schwarz2018progress}
Jonathan Schwarz, Wojciech Czarnecki, Jelena Luketina, Agnieszka
  Grabska-Barwinska, Yee~Whye Teh, Razvan Pascanu, and Raia Hadsell.
\newblock Progress \& compress: A scalable framework for continual learning.
\newblock In {\em International Conference on Machine Learning}, pages
  4528--4537. PMLR, 2018.

\bibitem{shu2020feature}
Chang Shu, Kun Yu, Zhixiang Duan, and Kuiyuan Yang.
\newblock Feature-metric loss for self-supervised learning of depth and
  egomotion.
\newblock In {\em Computer Vision--ECCV 2020: 16th European Conference,
  Glasgow, UK, August 23--28, 2020, Proceedings, Part XIX}, pages 572--588.
  Springer, 2020.

\bibitem{silberman2012indoor}
Nathan Silberman, Derek Hoiem, Pushmeet Kohli, and Rob Fergus.
\newblock Indoor segmentation and support inference from rgbd images.
\newblock {\em ECCV (5)}, 7576:746--760, 2012.

\bibitem{van2022three}
Gido~M van~de Ven, Tinne Tuytelaars, and Andreas~S Tolias.
\newblock Three types of incremental learning.
\newblock {\em Nature Machine Intelligence}, 4(12):1185--1197, 2022.

\bibitem{DBLP:conf/visapp/VarmaCZA22}
Arnav Varma, Hemang Chawla, Bahram Zonooz, and Elahe Arani.
\newblock Transformers in self-supervised monocular depth estimation with
  unknown camera intrinsics.
\newblock In {\em Proceedings of the 17th International Joint Conference on
  Computer Vision, Imaging and Computer Graphics Theory and Applications,
  {VISIGRAPP} 2022, Volume 4: VISAPP}, pages 758--769. {SCITEPRESS}, 2022.

\bibitem{vitter1985random}
Jeffrey~S Vitter.
\newblock Random sampling with a reservoir.
\newblock {\em ACM Transactions on Mathematical Software (TOMS)}, 1985.

\bibitem{voedisch2022continual}
Niclas V{\"o}disch, Daniele Cattaneo, Wolfram Burgard, and Abhinav Valada.
\newblock Continual slam: Beyond lifelong simultaneous localization and mapping
  through continual learning.
\newblock In {\em Proceedings of the International Symposium on Robotics
  Research (ISRR)}, 2022.

\bibitem{wang2018learning}
Chaoyang Wang, Jos{\'e}~Miguel Buenaposada, Rui Zhu, and Simon Lucey.
\newblock Learning depth from monocular videos using direct methods.
\newblock In {\em Proceedings of the IEEE conference on computer vision and
  pattern recognition}, pages 2022--2030, 2018.

\bibitem{Wang_2021_ICCV}
Lijun Wang, Yifan Wang, Linzhao Wang, Yunlong Zhan, Ying Wang, and Huchuan Lu.
\newblock Can scale-consistent monocular depth be learned in a self-supervised
  scale-invariant manner?
\newblock In {\em Proceedings of the IEEE/CVF International Conference on
  Computer Vision (ICCV)}, pages 12727--12736, October 2021.

\bibitem{wang2004image}
Zhou Wang, Alan~C Bovik, Hamid~R Sheikh, and Eero~P Simoncelli.
\newblock Image quality assessment: from error visibility to structural
  similarity.
\newblock {\em IEEE transactions on image processing}, 13(4):600--612, 2004.

\bibitem{yang2020d3vo}
Nan Yang, Lukas~von Stumberg, Rui Wang, and Daniel Cremers.
\newblock D3vo: Deep depth, deep pose and deep uncertainty for monocular visual
  odometry.
\newblock In {\em Proceedings of the IEEE/CVF conference on computer vision and
  pattern recognition}, pages 1281--1292, 2020.

\bibitem{yoon2017lifelong}
Jaehong Yoon, Eunho Yang, Jeongtae Lee, and Sung~Ju Hwang.
\newblock Lifelong learning with dynamically expandable networks.
\newblock {\em arXiv preprint arXiv:1708.01547}, 2017.

\bibitem{zenke2017continual}
Friedemann Zenke, Ben Poole, and Surya Ganguli.
\newblock Continual learning through synaptic intelligence.
\newblock In {\em International Conference on Machine Learning}, pages
  3987--3995. PMLR, 2017.

\bibitem{zhang2020online}
Zhenyu Zhang, Stephane Lathuiliere, Elisa Ricci, Nicu Sebe, Yan Yan, and Jian
  Yang.
\newblock Online depth learning against forgetting in monocular videos.
\newblock In {\em Proceedings of the IEEE/CVF Conference on Computer Vision and
  Pattern Recognition}, pages 4494--4503, 2020.

\end{thebibliography}
}

\newpage
\clearpage
\appendix
\resetlinenumber
\setcounter{page}{1}
\setcounter{table}{0}
\renewcommand{\thetable}{S\arabic{table}}

\setcounter{figure}{0}
\renewcommand{\thefigure}{S\arabic{figure}}

\section*{Supplementary Material}
\label{sec:Appendix}
\begin{table*}[t]
\centering
\small
\begin{tabular}{c|c|ccc|ccc|ccc}
\hline
\multirow{1}{*}{\textbf{Method}} & \multirow{2}{*}{\textbf{Buffer Size}} & \multicolumn{3}{c|}{$\mathbf{\mu_\textbf{final}}$} & \multicolumn{3}{c|}{$\mathbf{\mu_\textbf{overall}}$} & \multicolumn{3}{c}{\textbf{SPTO}} 
\\ \cline{3-11} 
 & & {\cellcolor{red!20}abs\_rel$\downarrow$} & {\cellcolor{red!20}RMSE$\downarrow$} & {\cellcolor{blue!20}a1$\uparrow$} & {\cellcolor{red!20}abs\_rel$\downarrow$}  & {\cellcolor{red!20}RMSE$\downarrow$}  & {\cellcolor{blue!20}a1$\uparrow$}  & {\cellcolor{red!20}abs\_rel$\downarrow$} & {\cellcolor{red!20}RMSE$\downarrow$} & {\cellcolor{blue!20}a1$\uparrow$} 
 \\ \hline \hline
\multirow{1}{*}{NCT}
& -- & 0.512  & 12.930  & 0.318  & 0.364  & 11.008  & 0.525  & 0.328  & 9.542  & 0.350 \\
\hline 
\multirow{3}{*}{} 
& 50 & 0.303 & 8.595 & 0.543 & 0.255 & 8.618 & 0.637 & 0.255 & 8.266 & 0.604 \\ 
{MonoDepthCL} & 200 & 0.260 & 7.544 & 0.633 & 0.268 & 8.561 & 0.655 & 0.255 & 8.160 & 0.664 \\
{} & 500 & 0.225 & 6.795 & 0.672 & 0.209 & 7.592 & 0.704 & 0.208 & 7.419 & 0.704 \\ \hline 
\end{tabular}
\caption{Performance on the CUDE framework for multiple sizes of the memory buffer, when the only short depth range task is learned last.}
\label{tab:appendix_depth_range}
\end{table*}

\begin{figure*}[h]
\centering
\includegraphics[width=.95\textwidth]{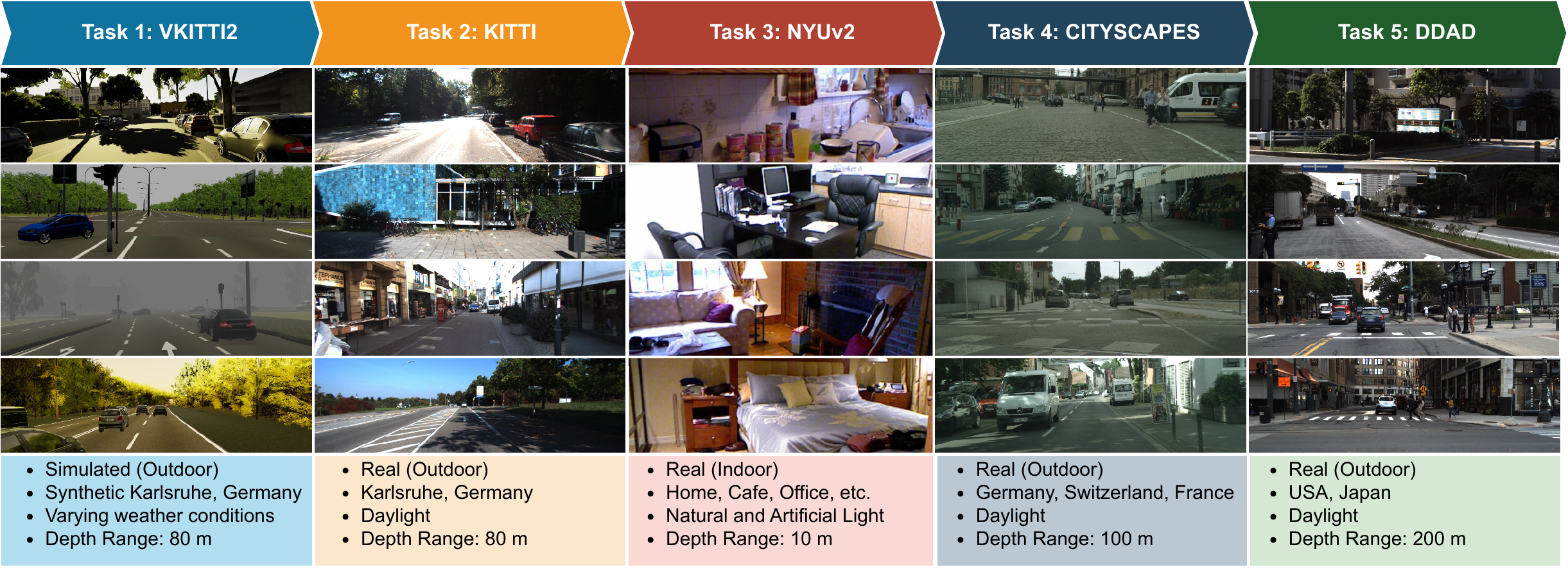}
\caption{Extended Continual Unsupervised Depth Estimation (CUDE) framework for 5 tasks.}
\label{fig:framework-5}
\end{figure*}

\section{Architecture and Hyperparameters}
\label{subsec:arch}
We train the baseline depth estimation model as described in Section~\ref{sec:method_depth_estimation} with a ResNet-18 backbone. The networks are implemented in PyTorch and trained on a Tesla V100 GPU with a batch size of 8 and a rehearsal batch size of 8. To control for factors such as dataset size and image size, we fix the training set size to 12,000 images and the test set size to 600 images for each task. The images are center-cropped to maintain the same aspect ratio as in the KITTI dataset and resized to a common width and height of 640 and 192 pixels, respectively. Thus, a total of 48000 samples are used for training with task identities assumed to be known and 2400 samples for testing with task identities assumed to be \textit{un}known. 
We establish a baseline where the four tasks are trained sequentially for 20 epochs each with an Adam optimizer and an initial learning rate of 1e$^{-4}$ for each task. The learning rate is decayed by a factor of 10 after 15 epochs for the respective task. Additionally, we train the four tasks jointly for a total of 20 epochs and an initial learning rate of 1e$^{-4 }$ which is decayed by a factor of 10 after 15 epochs. Finally, we apply a weight of $\beta=0.1$ for the spatiotemporal consistency loss with $\rho=0.85$ (Equation~\ref{eq:stc_terms}), and update the context model at a frequency of $0.05$ with $\alpha=0.999$. The results of MonoDepthCL are shown for the working model.

\section{Impact of task order on depth ranges}
\label{sec:appendix_task_order}

We pick our task order of VKITTI$\rightarrow$ KITTI$\rightarrow$ NYUv2$\rightarrow$ Cityscapes from the 24 possible combinations for reasons outlined in Section~\ref{sec:Framework} including covering all domain shifts such as indoor-to-outdoor, outdoor-to-indoor, and real-to-sim. However, NYUv2 has a short range depth ($10m$), whereas the remaining tasks, i.e. datasets have longer ($80m$-$100m$) range depths. This raises an interesting question - if NYUv2, i.e. short range depth estimation is the last task performed with no return to a longer depth range task such as Cityscapes, does our method still remember to perform a long range depth estimation?

To test this, we swap the order of appearance of NYUv2 and Cityscapes in our CUDE framework, and train MonoDepthCL on this swapped sequence. 
Table~\ref{tab:appendix_depth_range} shows that MonoDepthCL still outperforms NCT on all metrics at different buffer sizes by a large margin on this sequence.
Therefore, MonoDepthCL still remembers to perform long range depth estimation right after learning a short range depth estimation task, compared to NCT which undergoes catastrophic forgetting of long range depth estimation.

\section{Unsupervised Depth Estimation}
\label{sec:appendix_method_depth_estimation}

Here, we provide some of the details of the unsupervised monocular depth estimation method (see Section~\ref{sec:method_depth_estimation}).

\paragraph{Depth network:} The depth network $f_D$ parameterized by $\theta_D$ predicts inverse depths at four resolutions as follows:
\begin{equation}
    \label{eq:depth_fwd}
        D^{-1}_1, D^{-1}_2, D^{-1}_3, D^{-1}_4 = f_D(I_t; \theta_D).
\end{equation}

\paragraph{Ego-motion network:} The ego-motion network $f_E$ parameterized by $\theta_E$ predicts the relative rotation $R_{s \leftarrow t}$ and translation $T_{s \leftarrow t}$ between each source-target image pair concatenated along the channel dimension as follows:
\begin{equation}
    \label{eq:pose_fwd}
         R^j_{s \leftarrow t}, T^j_{s \leftarrow t} = f_E(I^j_s, I_t; \theta_E), 
\end{equation}

\paragraph{Perspective projection:}
 For each source image $I^j_{s}$ when warping to the $i^{th}$ upsampled target image using the $i^{th}$ depth prediction; 
\begin{equation}
\label{eq:perspective_model}
    p^j_{s} \sim K{R}^j_{s \leftarrow t} {D_i}[p_{i,t}]K^{-1}p_{i,t} + K{T}^j_{s \leftarrow t},
\end{equation}
where $K$ is the camera intrinsics matrix, and $p^j_s$ and $p_{i,t}$ refer to the pixel locations in $j^{\text{th}}$ source and target images, respectively. Then, we use bilinear interpolation to obtain the value of the warped image $I^j_s$ at each location $p_{i,t}$.

\paragraph{Photometric error:}
The appearance based per-pixel photometric error between the original \textit{target} image and the synthesized target images from $n_s$ \textit{source} images for the $i^{th}$ prediction (Equation~\ref{eq:depth_fwd}) is defined as follows:
\begin{equation}
\label{eq:photometric_loss}
\begin{split}
    \mathcal{P}^j_i &= \frac{\rho}{2}(1 - SSIM(I_t, \hat{I}^j_{i,t})) + (1 - \rho)\lVert I_t - \hat{I}^j_{i, t} \rVert_1, \\
    \mathcal{P}_i &= \min_j \mathcal{P}^j_i, j = 1, 2,...n_s. \\
\end{split}
\end{equation}
This per-pixel minimum serves to deal with out-of-view pixels and occlusion, such that only the source for which the synthesis is most accurate contributes to the error term. As discussed earlier in Section~\ref{sec:method_depth_estimation}, the loss is masked to counteract the impact of temporally stationary pixels.

\paragraph{Smoothness loss:}
The per-pixel edge-aware \textit{smoothness loss} for the $i^{th}$ prediction (Equation~\ref{eq:depth_fwd}) is defined as follows:
\begin{equation}
\label{eq:smoothness_loss}
    \mathcal{S}_i = \left\lvert \partial_x \frac{D^{-1}_i}{\mathbb{E}_{p_t}[D^{-1}_i]} \right \rvert \mathrm{e}^{-\lvert\partial_x I_t\rvert} 
+
\left\lvert \partial_y \frac{D^{-1}_i}{\mathbb{E}_{p_t}[D^{-1}_i]} \right \rvert \mathrm{e}^{-\lvert\partial_y I_t\rvert},
\end{equation}
where the expectation $\mathbb{E}$ of inverse depth predictions are computed across all target pixels~\cite{wang2018learning}.

\paragraph{Total task loss :}
The total combined training loss across all $4$ predictions for unsupervised depth estimation is:
\begin{equation}
\label{eq:depth_loss}
L_{depth} = \frac{1}{4HW} \sum_{i=1}^4 \sum_{p_t \in I_t} \mathcal{P}_i[p_t] + \frac{\lambda}{2^ {i - 1}} \mathcal{S}_i[p_t].
\end{equation}



\end{document}